%% file: main.tex
\pdfoutput=1

\documentclass[11pt]{article}

\usepackage[]{acl}

\usepackage{times}
\usepackage{latexsym}
\usepackage{changes}

\usepackage[T1]{fontenc}

\usepackage[utf8]{inputenc}

\usepackage{microtype}

\usepackage{inconsolata}

\usepackage{graphicx}

\usepackage{xcolor}

\usepackage{multirow}
\usepackage{arydshln}
\usepackage{enumitem}
\usepackage{xspace}
\usepackage{pdfpages}

\usepackage{listings}
\usepackage{caption}
\usepackage{fancyvrb}

\usepackage{booktabs}     %
\usepackage{subcaption}   %

\lstdefinestyle{promptstyle}{
    basicstyle=\ttfamily\small,
    breaklines=true,
    frame=single,
    backgroundcolor=\color{gray!10},
    numbers=left,
    numberstyle=\tiny,
    numbersep=5pt,
    keywordstyle=\bfseries,
    xleftmargin=10pt,
    xrightmargin=10pt
}

\newcommand{\dataset}{\textsc{CaseSumm}\xspace}
\newcommand{\casesummsize}{25.6K\xspace}

\newif\ifcomments

\commentstrue

\ifcomments
  \newcommand{\chenhao}[1]{\textcolor{magenta}{\textsc{#1 ---CT}}}
\else
  \newcommand{\chenhao}[1]{}
\fi

\ifcomments
  \newcommand{\mourad}[1]{\textcolor{blue}{\textsc{#1 ---MH}}}
\else
  \newcommand{\mourad}[1]{}
\fi

\title{\dataset: A Large-Scale Dataset for Long-Context Summarization from U.S. Supreme Court Opinions}

\author{Mourad Heddaya \\
University of Chicago\\
\texttt{mourad@uchicago.edu} \\
\And
Kyle MacMillan \\
University of Chicago\\
\texttt{macmillan@uchicago.edu}\\
\AND
Anup Malani \\
University of Chicago Law School\\
\texttt{amalani@uchicago.edu} \\
\And
Hongyuan Mei \\
TTIC\\
\texttt{hongyuan@ttic.edu}\\
\And
Chenhao Tan \\
University of Chicago\\
\texttt{chenhao@uchicago.edu}\\
}

\begin{document}
\maketitle

\input{sections/abstract}
\input{sections/introduction}

\input{sections/related}

\input{sections/dataset}

\input{sections/experiments}

\input{sections/results}

\input{sections/conclusion}

\input{sections/limitations}

\section*{Acknowledgments}
This work is supported in part by NSF grants IIS-2302785, an award from the
Sloan Foundation, and gifts from Google and Amazon.

\bibliography{custom}

\clearpage
\appendix
\input{sections/appendix}

\end{document}

%% file: sections/abstract.tex
\begin{abstract}
This paper introduces \dataset, a novel dataset for long-context summarization in the legal domain that addresses the need for longer and more complex datasets for summarization evaluation. We collect \casesummsize U.S. Supreme Court (SCOTUS) opinions and their official summaries, known as "syllabuses." Our dataset is the largest open legal case summarization dataset, and is the first to include summaries of SCOTUS decisions dating back to 1815.

We also present a comprehensive evaluation of LLM-generated summaries using both automatic metrics and expert human evaluation, revealing discrepancies between these assessment methods. Our evaluation shows Mistral 7b, a smaller open-source model, outperforms larger models on most automatic metrics and successfully generates syllabus-like summaries. In contrast, human expert annotators indicate that Mistral summaries contain hallucinations. The annotators consistently rank GPT-4 summaries as clearer and exhibiting greater sensitivity and specificity. Further, we find that LLM-based evaluations are not more correlated with human evaluations than traditional automatic metrics. Furthermore, our analysis identifies specific hallucinations in generated summaries, including precedent citation errors and misrepresentations of case facts. These findings demonstrate the limitations of current automatic evaluation methods for legal summarization and highlight the critical role of human evaluation in assessing summary quality, particularly in complex, high-stakes domains.

\dataset is available on \href{https://huggingface.co/datasets/ChicagoHAI/CaseSumm}{HuggingFace}.\footnote{\href{https://huggingface.co/datasets/ChicagoHAI/CaseSumm}{https://huggingface.co/datasets/ChicagoHAI/CaseSumm}}
\end{abstract}

%% file: sections/introduction.tex
\section{Introduction}

Although large language models (LLMs) are claimed to handle long contexts~\citep{openai2024gpt4technicalreport, bubeck2023sparksartificialgeneralintelligence, claude}, including 
summarizing very long inputs, how well they perform long-context summarization is an open question.

Evaluating long-context summarization is challenging for several reasons.
First, human ground-truth summaries are often not available~\citep{cao2024characterizingmultimodallongformsummarization, chang2024booookscore}. Moreover, it's unclear whether we should trust human abilities to even create ground-truth summaries.
Second, what makes a good summary in one setting may not generalize to other settings.
For example, what's relevant in a legal text is different than what's relevant in a novel.
Lastly, identifying salient information in complex domains often requires expertise.

We address these challenges by introducing a new dataset where ``ground-truth'' summaries are available and conducting a comprehensive human evaluation to benchmark existing models.
In particular, we build \dataset, a legal case summarization dataset consisting of \casesummsize U.S. Supreme Court cases and their official summaries, called syllabuses.  
Syllabuses are written by an attorney employed by the Court and approved by the Justices. The syllabus is therefore the gold standard for summarizing majority opinions, and ideal for evaluating other summaries of the opinion. We obtain the opinions from Public Resource Org's archive\footnote{\href{https://public.resource.org/}{https://public.resource.org/}} and extract syllabuses from the official opinions published in the U.S. Reporter and hosted by the Library of Congress. Our dataset is at least ~25\% larger, covers 3 times as many years (1815-2019), and is publicly available with fewer copyright restrictions than similar legal datasets \citep{fang2023super, indianlegalcorpus}, representing a rich resource for the research community.

Beyond the legal domain, several datasets have been introduced to improve evaluation of long-context summarization~\citep{kryściński2022booksumcollectiondatasetslongform, sharma-etal-2019-bigpatent, billsum, govreport}. \dataset{} continues the trend of larger datasets with both longer source and summary texts, where the summaries represent high quality ground-truths. Unlike prior work, however, our dataset spans over two centuries, demonstrating unique variation in the lengths and compression rates of summaries, while also reflecting a high-stakes and useful domain for summarization.

To highlight the opportunities and challenges of our dataset, we present both automatic and human expert evaluations of LLM-generated summaries of SCOTUS opinions and include two ``control'' human-written summaries from Westlaw and Oyez. According to both human and automatic metrics, fine-tuning Mistral successfully guides the model to more accurately mimic the official syllabuses and reflect the lexical and semantic content within them, than other much larger models.

However, overall we observe that automatic metrics do not correlate well with human judgments, and that LLM-based evaluation does not do better. We find that syllabuses and fine-tuned Mistral summaries perform highly on automatic evaluation but rank lower according to human evaluators, whereas GPT-4 is reliably ranked highly in human evaluation despite only average performance on automated metrics. Furthermore, GPT-4-generated summaries often outperform human-written ones, including official syllabuses, but not on factual correctness. These findings challenge the notion of human-written ground-truth summaries.

Finally, we conduct an error analysis of hallucinations in GPT-4- and Mistral- generated summaries and identify factual errors ranging from precedent citation errors to misrepresentations of the facts of the case and procedural history as recounted in the source opinions.

In sum, we make the following contributions:

\begin{itemize}[leftmargin=*]
    \item We introduce a new large-scale dataset for long-context summarization in the legal domain, consisting of \casesummsize U.S. Supreme Court cases and their official syllabuses from 1815-2019.
    \item We present a comprehensive evaluation of LLM-generated summaries using both automatic metrics and expert human evaluation, revealing discrepancies between these assessment methods.
    \item We provide a comparative analysis of summaries generated by fine-tuned models and larger, general-purpose models, offering insights into their relative strengths and weaknesses in legal summarization tasks.
\end{itemize}

%% file: sections/related.tex
\section{Related Work}

\paragraph{Evaluation for summarization.}
ROUGE \citep{lin-2004-rouge} has been the dominant summarization metric, despite criticism of its high lexical dependence \citep{schluter2017limits, cohan-goharian-2016-revisiting}. Newer metrics like BERTScore \cite{zhang2019bertscore} and BARTScore \citep{yuan2021bartscore} aim to capture semantic similarities. However, automatic metrics often don't correlate well with human judgments \citep{yuan2021bartscore, summeval-fabri, bhandari-etal-2020-evaluating}. We focus on high-stakes long-context summarization, showing the need for better metrics persists despite LLM progress. \citet{chang2024booookscore} extended LLM-based evaluation to book-length summaries, but this approach doesn't consider how experts weigh the importance of including or omitting certain information in a summary, while also being slow and costly.
\citet{cao2024characterizingmultimodallongformsummarization} developed a framework for characterizing LLM summaries of financial documents. Our work extends this research by evaluating and comparing model- and human-generated summaries in the legal domain. Addressing factual discrepancies in model-generated summaries, recent work has developed automatic methods for evaluating faithfulness in summarization \citep{krishna-etal-2023-longeval, chang2024booookscore, falke-etal-2019-ranking, summac, wang-etal-2020-asking, fabbri-etal-2022-qafacteval}.

\paragraph{NLP and summarization in the Legal Domain.}
Natural language processing has been applied to various legal tasks, including summarization~\citep{bauer2023legal}, discovery~\citep{zou2020towards}, redaction~\citep{garat2022automatic}, case outcome prediction~\citep{medvedeva2023rethinking,cui2023survey}, and Bar Exam performance~\citep{katz2023gpt}. For comprehensive surveys of NLP in the legal domain, see \citet{katz2023natural} and \citet{kapoor2024promises}.

\paragraph{Datasets in the legal domain.}
Our dataset is unique in providing U.S. Supreme Court opinions with syllabuses, unlike other datasets that lack syllabuses~\citep{chalkidis-etal-2022-lexglue, henderson2022pile} or provide only ancillary data~\citep{scdb2024}. 
\citet{fang2023super} present Super-SCOTUS, a dataset of Supreme Court documents, including a subset of syllabus (scraped from online websites and not validated) and opinion pairs and highlight its contribution to political and social science research. 
In contrast to Super-SCOTUS, our \dataset{} dataset consists of cleaned and carefully trimmed opinion and syllabus pairs. For each decision PDF, we identify and extract the syllabus and majority opinion, which syllabuses are intended to summarize. We remove headers and concurrent and dissenting opinions, while properly including footnotes.  As Table \ref{tab:other_dataset_comp} shows, \dataset is a strict super-set of Super-SCOTUS with descriptive statistics that reflect our improved data processing pipeline. \dataset extends further back to 1815 and, by being extracted directly from source opinions, provides the community with a readily available summarization resource with fewer copyright restrictions.

\

%% file: sections/dataset.tex
\section{Dataset}

When the Supreme Court resolves a case, it publishes a majority "opinion" announcing the outcome and reasoning for their decision. The Court will also disseminate a summary of the opinion called the "syllabus", which is written by an attorney employed by the Court and approved by the Justices. The syllabus must include the main elements of the opinion: the facts of the case, the procedural history, the legal question to be decided, and the answer to that question. Accurately summarizing each of these sections requires (1) understand sophisticated legal reasoning and (2) identify the most salient aspects of the case.

As one of the longest standing institutions in U.S. history, the Supreme Court has published thousands of opinions and syllabuses over the past 200 years. Looking at cases between 1815 and 2019, we collect \casesummsize pairs of opinions and syllabuses for our dataset, to be available under a CC BY-NC 4.0 license.

\paragraph{Dataset construction}

We compile our dataset from multiple sources. Opinions published in U.S. Reports Volume 15-546 (years 1815-2005) and Volumes 546-591 (2005 through \textit{Trump v. Vance} (2019)) are obtained from Public Resource Org's online archive~\citep{pro2024} and the Super-SCOTUS data set~\citep{fang2023super}, respectively. We extract 
syllabuses from PDFs of the opinions hosted on the Library of Congress's website~\citep{loc2024}. 

Extracting syllabuses from the original PDFs is challenging for several reasons. First, identifying the start and end of the syllabus is complicated because the formatting and style of SCOTUS decisions have changed over time. Low quality scans of 19th and 20th century documents make the extraction task even more difficult. Together, these issues constrain the kinds of rules, or signals, we can leverage to reconstruct the structure of the text in the PDFs, requiring us to identify alternatives. For example, while syllabuses have a smaller font-size than the rest of the decision and would be a straight-forward heuristic to leverage, this information is often incorrectly encoded in OCR data.

To ensure accurate syllabus extraction, we process the PDFs in multiple ways. First, we design a set of regular expressions to identify the start of the syllabus, providing coverage of decisions with different styles. Then, we develop an algorithm based on open-source computer vision software \citep{opencv-library} to identify continuous lines, allowing us to distinguish footers from the main text of a page. Finally, we take advantage of differences in line density, a measure that is more robust to OCR and scan quality, combined with regular expressions to determine when the syllabus ends. 

Since we build a new dataset, there are no accessible ground-truths to automatically evaluate our technique for extracting syllabuses from PDFs. Instead, we randomly sample 100 cases and manually evaluate the extracted syllabus by comparing them to the original PDFs. We find that 96 of the 100 are perfect extractions while the remaining 4 syllabuses are partially truncated. These results highlight the quality of our dataset as a rich resource for long-context summarization.

\paragraph{Descriptive statistics}\label{sec:desc-stats}

\begin{figure*}
    \centering
    \includegraphics[width=1\linewidth]{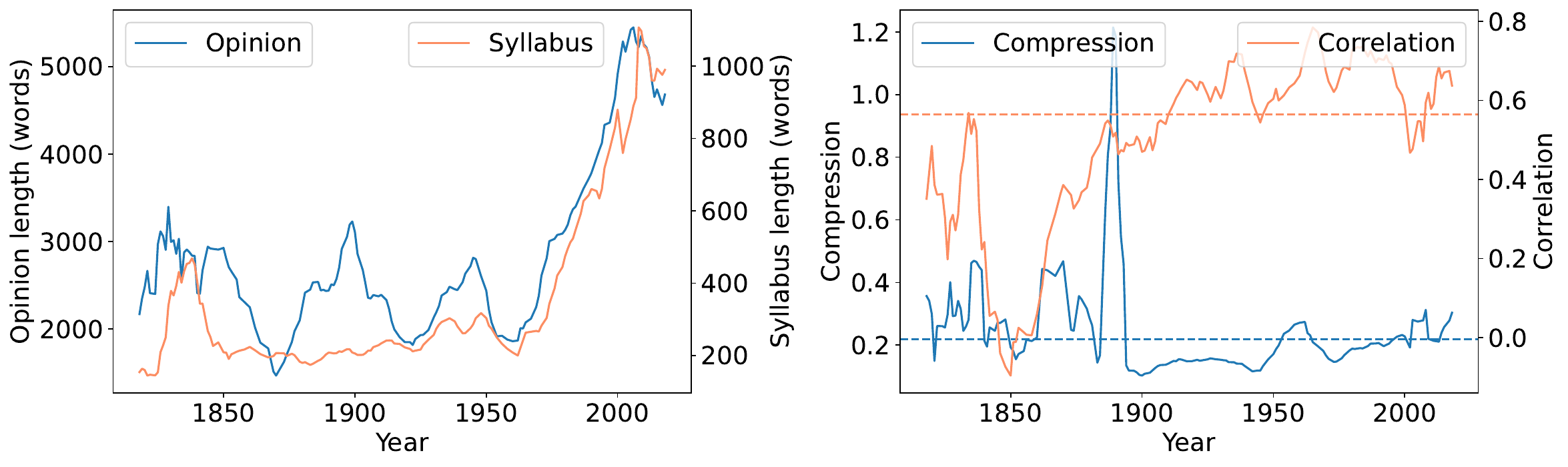}
    \caption{Opinion and syllabus lengths, compression rates by syllabuses, and correlations between opinion and syllabus lengths, 1815-2019.  Dashed blue and orange lines give average compression rate and correlation. Lines are smoothed with 5-year moving-average.}
    \label{fig:dataset_stats}
\end{figure*}

\input{assets/tables/dataset_comp}

To demonstrate the value of our dataset as a resource for abstractive summarization, we compare the lengths of the opinions and their syllabuses. 
The average Supreme Court opinion is 2,612 words long. The average syllabus is 314 words long, about 21.8\% the length of the opinion it is summarizing. Figure \ref{fig:dataset_stats} shows lengths have risen over time. Since 1980, opinions and syllabuses average 4,151 and 731 words, respectively, nearly double the average for the entire 1815-2019 period.  Although compression rates, defined as the ratio of words in a syllabus to words in an opinion have been relatively stable over time, averaging 21.8\% from 1815-2019, 
the Pearson correlation between the length of an opinion and its syllabus, while variable, has increased over time.  Whereas this correlation was just 0.46 before 1920, it has been 0.68 since then.  Given the changes in opinion and syllabus lengths and in the correlation between syllabus and opinion lengths, this data set is a valuable resource for modeling and evaluating expert summaries, especially in the legal domain.

%% file: assets/tables/dataset_comp.tex
\setlength\dashlinedash{0.2pt}
\setlength\dashlinegap{1.5pt}
\setlength\dashlinedash{0.2pt}
\setlength\dashlinegap{1.5pt}
\begin{table}
\centering
\small
\resizebox{\columnwidth}{!}{%
\begin{tabular}{lrrr}
\toprule
\addlinespace[0.3em]
\textbf{Dataset} & \textbf{\# Docs}. & \multicolumn{2}{c}{\textbf{\# Words}} \\
& & \textbf{Source} & \textbf{Summary} \\
\midrule
Super-SCOTUS (1955-2019) & 6.6k & 9.3k & 791 \\
\addlinespace[0.2em]
BillSum (1993-2018) & 22.2k & 1.8k & 208 \\
\addlinespace[0.2em]
GovReport* & 18.5k & 9.4k & 553 \\ 
\addlinespace[0.2em]
Multi-LexSum** & 4.5k & 75.5k & 647 \\
\addlinespace[0.15em]
\hdashline\noalign{\vskip 0.5ex} 
\addlinespace[0.15em]
Oyez (1955-2012)*** & 622 & 4.8k & 356 \\
\addlinespace[0.2em]
Westlaw (1956-2011) & 156 & 4.5k & 143 \\
\addlinespace[0.2em]
\textbf{\dataset} (1815-2019) & 25.6k & 2.6k & 314 \\
\hspace{3mm} \textbf{\dataset} (1955-2019) & 7.2k & 4.9k & 745 \\
\hspace{3mm} \textbf{\dataset} (1815-1955) & 18.4k & 3.4k & 289 \\
\bottomrule
\end{tabular}
}
\caption{Comparison of \dataset and related long-context summarization datasets in the legal domain. *GovReport does not report the range of years covered. **Multi-LexSum is a multi-document summarization dataset. ***Oyez summaries are a subset of SuperSCOTUS.}
\label{tab:other_dataset_comp}
\end{table}

%% file: sections/experiments.tex
\section{Experiment Setup}

In this section, we introduce our summarization task setup and evaluation strategies.

\subsection{Data and Modeling}

\paragraph{Data preprocessing and splits.}

We use syllabuses as a supervision signal in our summarization modeling experiment and as reference summaries for evaluating the human and model-generated summaries. 

As discussed in \S\ref{sec:desc-stats}, the substance and style of syllabuses have changed over time. Therefore, the supervision signal has changed over time. The motivating use case in our summarization task is a legal professional conduction research. For such a professional, while concision has value, comprehensiveness is more valuable. By manually studying summaries, we determine that more comprehensive syllabuses begin with a summary of the facts of the case, followed by a new section---marked by the text ``Held:''---containing details about the issues, analyses, and conclusions that the opinion commented on. Modern syllabuses 
consistently adhere to this structure.

Therefore, we filter our dataset to include only opinion/syllabus pairs where the syllabus contains the pattern ``Held:''. We call this subset of the dataset ``structured".  We find that the length of structured syllabuses is more strongly correlated with the length of their respective opinions ($r=0.65$) than the length of unstructured syllabuses is with the length of their opinions ($r=0.46$). Furthermore, structured syllabuses are on average 2.5x longer than the unstructured syllabuses.  Overall, the structured dataset contains 6,683 case/syllabus pairs. We split these into a training set ($n$=5,455), validation set ($n$=606), and test set ($n$=622).

\paragraph{Modeling.}

We pursue and test two approaches for completing our legal case summarization task. The first approach is zero-shot prompting with proprietary and with open-source LLMs. The propriety LLM we employ is GPT-4 Turbo (\verb|gpt-4-1106-preview|)~\citep{openai2024gpt4technicalreport}. The open-source LLM we employ is Mistral 7b Instruct (\verb|Mistral-7b-Instruct-v0.1|)~\citep{jiang2023mistral}. The opinions in our dataset have 4,983 tokens on average, and the syllabuses average 755 tokens. 
The second approach is instruction fine-tuning~\citep{wei2021instruction} the open-source model, Mistral 7b Instruct, using the syllabuses in our training data set.   
We will refer to models used in a zero-shot setting by model name: Mistral Base and GPT-4, and to the fine-tuned Mistral model as Mistral FT.   

For Mistral in both settings, we design a prompt following best practices suggested by its authors.\footnote{\href{https://huggingface.co/mistralai/Mistral-7B-Instruct-v0.1}{https://huggingface.co/mistralai/Mistral-7B-Instruct-v0.1}} For GPT-4, we optimize prompt-selection using DSPy \citep{khattab2023dspy} with 10 opinion/syllabus pairs from the training set and ROUGE-2 as the optimization metric.

For fine-tuning, our input consists of a short instruction, the opinion, and the syllabus. We do standard auto-regressive language modeling but only backpropagate the language modeling loss for the syllabus. We use LoRA-based Parameter-Efficient Fine-Tuning (PEFT)~\citep{hu2021lora} to train a subset of the parameters. 

We include additional implementation details in Appendix \ref{sec:summary_gen_appendix}.

\subsection{Evaluation strategies}

\input{assets/tables/auto-metrics/all_auto_syllabus}

\paragraph{``Control group'' summaries.}  
We benchmark our three machine-generated summaries (Mistral Base, Mistral FT, and GPT-4 Turbo) along with two additional human-generated sources for purposes of having a control group of human-written summaries not explicitly intended to mimic syllabuses. First, we collect public Oyez summaries from the Super-SCOTUS dataset~\citep{fang2023super}. Oyez summaries are composed of three sections: Facts of the Case, Question, and Conclusion. Second, we collect Westlaw's commercial summaries of cases via their online interface.\footnote{We obtain these manually to avoid legal risks under our Westlaw subscription license.} Because manual download is slow, our sample size for Westlaw downloads was smaller: whereas our test set has 622 instances of model-generated summaries and Oyez summaries, we have 156 Westlaw summaries.\footnote{We initially included summaries from Justia, another publicly available legal resource, as a human baseline but, after manually inspecting 5 randomly sample summaries, we determine that they were largely derivative of the Court syllabuses and copied significant quantities of text from them. This was further validated by finding that Justia summaries achieved 0.97 ROUGE-1 score, which is exceedingly alike in a long-form summarization task such as this.}

\paragraph{Automatic Evaluation.}
Following recent work on summarization \citep{Koh-2022}, we use ROUGE and BERTScore \citep{lin-2004-rouge, zhang2019bertscore} as our automated metrics for evaluating generated summaries against the reference syllabuses. With this, we assess the \textit{relevance} of the summaries. We breakdown each of the metrics by their precision, recall, and F1-score, highlighting how models balance trade-offs between coverage and concision. We also experimented with BARTscore \citep{yuan2021bartscore} (see Appendix \ref{sec:bartscore-results}) but exclude it from our main analysis due to its sensitivity to whether text is in- or out-of- distribution relative to the scoring model. Since we compare Mistral after fine-tuning on syllabuses to models that were not fine-tuned, we expect unreliable results.

To further characterize the summaries, we compare the summaries based on \textit{compression rate}, defined as the number of words in a syllabus over the number of words in an opinion, and the \textit{correlation} between opinion lengths and summary lengths. We use compression as a measure of brevity and correlation as a measure of how responsive summaries are to changes in the amount of content in the opinions.

\paragraph{Human Evaluation.}

For the human evaluation, we recruited and paid\footnote{Participants were paid \$20/hr, \$4 more than RA minimum. See Appendix \ref{sec:consent} for instructions \& consent.} second- and third-year law students to read several opinions and 5 summaries of each opinion (Mistral FT\footnote{We exclude Mistral base from our comparisons because it has rather poor performance overall on automatic metrics, helping us reduce the cognitive load on our participants.}, GPT-4, official syllabus, Westlaw, and Oyez).  We asked students to rank each summary (from 1 to 5) on several metrics: did the summary contain all relevant information from the opinion (\textit{sensitivity}), did it exclude irrelevant information (\textit{specificity}), was the summary clear (\textit{clarity}), and did the summary have a style suggesting it was written by an experienced attorney? (\textit{style})  Finally, we asked students to report the number of facts in the summary that were false based on their reading of the opinion (\textit{error}).   Students were not told the source of each summary.\footnote{This evaluation was deemed exempt from IRB review by our institution's IRB (IRB24-0277).} See Appendix \ref{sec:human-eval-app} for additional details on the annotation interface and procedure.

In total, students read 57 opinions.  Our sample of opinions and summaries included 33 unique cases, and the median student read 5 cases.  Given that we ask students to rank opinions from 1 to 5 (implying a mean of 3 and variance of 2), our minimum detectable effect, with 95\% confidence and 80\% power, was 0.52 rank points.

\paragraph{Experimenting with LLM-based evaluations.}

Metrics like ROUGE and BERTScore provide a baseline for assessing lexical and semantic alignment between a candidate and reference text. However, they can miss deeper qualities that humans value in a good summary. LLMs offer a new way to evaluate summaries and to address some of these shortcomings \citep{liu2023gevalnlgevaluationusing, song-etal-2024-finesure}. Still, their results are variable and sensitive to their prompts. In this work, we study how well G-Eval \citep{liu2023gevalnlgevaluationusing}, a GPT-4-based evaluation tool, agrees with human ratings compared to ROUGE and BERTScore. This helps us understand whether G-Eval offers a better way to evaluate summaries when a reference is unavailable or when traditional metrics fall short. We test both the default implementation of G-Eval, as well as an adapted version to more closely reflect our human evaluation setup.

\paragraph{Correlation between automatic and human rankings.} 
In Section \ref{sec:human_results}, we discuss differences and similarities in how various evaluation methods, including G-Eval, correlate with human judgments. For each opinion, we convert the ROUGE, BERTScore, and G-Eval scores for the various candidate summaries into rankings. Then, we compute the Spearman correlation between each ranking and the human ranking, and average these correlations.

%% file: assets/tables/auto-metrics/all_auto_syllabus.tex
\setlength\dashlinedash{0.2pt}
\setlength\dashlinegap{1.5pt}

\begin{table*}[ht]
\centering
\small
\resizebox{\textwidth}{!}{%
\begin{tabular}{@{}lcccccccccccccccc@{}}
\toprule
   & \multicolumn{3}{c}{\textbf{ROUGE-1 $\uparrow$}} & \multicolumn{3}{c}{\textbf{ROUGE-2 $\uparrow$}} & \multicolumn{3}{c}{\textbf{ROUGE-L $\uparrow$}} & \multicolumn{3}{c}{\textbf{BERTScore $\uparrow$}} \\
 \cmidrule(lr){2-4} \cmidrule(lr){5-7} \cmidrule(lr){8-10} \cmidrule(lr){11-13}
\textbf{Method}   & P & R & F1 & P & R & F1 & P & R & F1 & P & R & F1 \\ \midrule
\addlinespace[0.3em]
GPT-4 Turbo         & \underline{71.2} & \underline{37.1} & \underline{45.1} & \underline{31.1} & \underline{15.5} & \underline{19.2} & 34.9 & 18.1 & \underline{21.9} & \textbf{67.4} & \underline{62.2} & \textbf{64.6} \\
\addlinespace[0.3em]
Mistral Base        & 64.3 & 13.4 & 20.0 & 23.4 & 4.6 & 7.0 & \underline{41.1} & 7.8 & 11.8 & 61.3 & 48.5 & 54.0 \\
\addlinespace[0.3em]
Mistral FT   & 63.3 & \textbf{43.1} & \textbf{48.1} & 30.1 & \textbf{20.5} & \textbf{23.0} & 34.9 & \textbf{23.6} & \textbf{26.4} & \underline{66.0} & \textbf{64.4} & \textbf{65.1} \\
\addlinespace[0.15em]
\hdashline\noalign{\vskip 0.5ex} 
\addlinespace[0.15em]
Oyez                & 64.0 & 35.1 & 41.6 & 28.5 & 15.0 & 18.1 & 34.4 & \underline{18.6} & \underline{22.1} & 64.2 & \underline{61.8} & \underline{62.9}  \\
\addlinespace[0.3em]
Westlaw             & \textbf{71.5} & 20.5 & 29.4 & \textbf{32.7} & 9.1 & 13.2 & \textbf{42.3} & 11.8 & 17.0 & \underline{65.0} & 55.7 & 59.9  \\
\bottomrule
\end{tabular}
}
\caption{Automatic evaluation of model-generated and human-written summaries, where official syllabuses are the reference summaries. Sample includes 622 Supreme Court cases. There are 622 observations on each type of summary except Westlaw, for which we only have 156 observations. For each metric, we report precision (P), recall (R), and F1-score (F1). For each metric, we \textbf{bold} the best score(s) and \underline{underline} the second best score(s).
}
\label{tab:summary-eval-syllabus}
\end{table*}

%% file: sections/results.tex
\section{Results}
Our results indicate consistencies and discrepancies in the outcomes of automatic and human evaluations. On the one hand, model-generated summaries largely outperform the control human-written summaries on automatic measures of relevance, while also matching or exceeding them in our human evaluation. On the other hand, automatic metrics prefer Mistral FT summaries over GPT-4 ones, whereas expert humans most commonly rank GPT-4 over Mistral FT. Furthermore, we show that all summaries are shorter than their reference syllabuses and do not correlate as strongly with opinion lengths. Despite this, humans prefer GPT-4 summaries, revealing that its summaries may represent a more desirable trade-off between concision and comprehensiveness.

\subsection{Automated Evaluation Favors Fine-tuned Mistral Summaries}

We start by looking at the results in Table \ref{tab:summary-eval-syllabus} of automatic evaluation between summaries and official syllabuses for the three generated summaries (Mistral Base, Mistral FT, and GPT-4) and for two human summaries. Overall, we find that fine-tuning Mistral is particularly effective at improving the recall scores across all the metrics: ROUGE recall scores increase by an average of 21 points, BERTScore recall by 15 points.
However, effects of fine-tuning on precision are weaker and more mixed. Perhaps fine-tuning sacrifices brevity for inclusion of more words in a syllabus, i.e., improves the sensitivity of summaries at a cost to specificity.

\paragraph{Control summaries help highlight effects of style differences on automatic metrics.}
By comparing against the two control human-written summaries, we can clearly see that Westlaw is an outlier.
While GPT-4 and Mistral FT scores mostly resemble Oyez, Westlaw's recall scores are particularly low, only surpassing Mistral base. This poor performance on recall, but strong performance on precision, may be a product of how short those summaries are.

\subsection{Summaries do not Scale with Opinion Length as much as Official Syllabuses}

\input{assets/tables/summary_describe}

A unique aspect of \dataset{} is that it includes SCOTUS cases dating back to more than two centuries ago. This breadth enables researchers to investigate summaries from many different angles. In this subsection, we characterize candidate summaries through the lens of length and compression and explore how these variables may affect summary quality over time.

\paragraph{Length \& Compression.}
In our dataset, both the opinion and syllabus lengths systematically co-vary across time. 
Table~\ref{tab:summary_describe} shows that syllabuses in our sample have an average compression rate of 17.6\%, meaning they tend to be about one-sixth the length of the original opinions. We find that in generated summaries, Mistral FT produces summaries closest in length to these syllabuses, even outperforming GPT-4, which was also prompt-optimized to mimic syllabuses. Westlaw produced the shortest summaries, followed by Mistral without fine-tuning.

Regarding the correlation between summary and opinion lengths, syllabuses demonstrate the strongest relationship with opinion lengths: doubling opinion length increases syllabus length by nearly 2/3. In contrast, Mistral FT summaries show a weaker correlation, with doubling opinion length increasing summary length by only 18\%. Westlaw summaries exhibit almost no correlation with opinion length, maintaining a consistent target length of approximately 150 words. These findings highlight our dataset as a rich resource for future work in investigating how automatic summarization methods may adapt to varying source document lengths, ensuring that all salient information is captured regardless of length.

\begin{figure}[t]
    \centering
    \includegraphics[width=\linewidth]{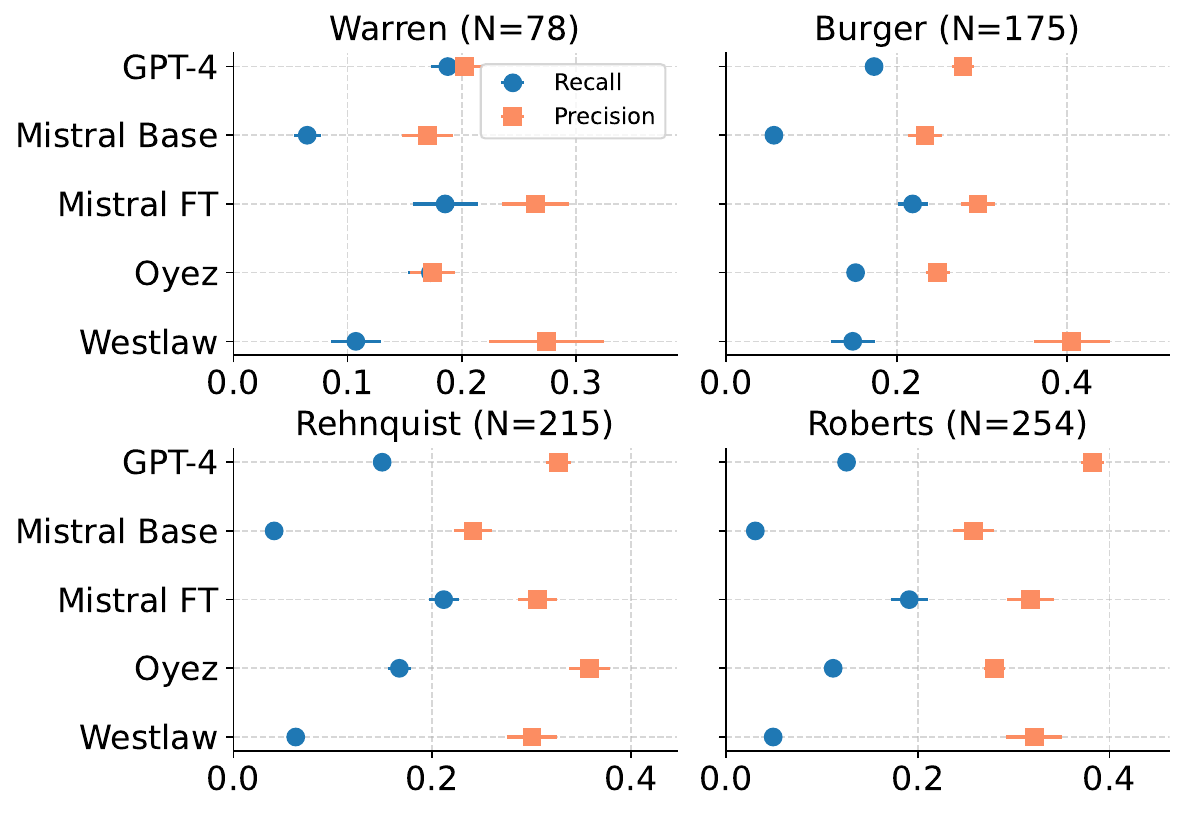}
    \caption{ROUGE-2 evaluation of model-generated and human summaries, by Chief Justice of SCOTUS when the opinion was written. Markers are means and whiskers are 95\% confidence intervals.}
    \label{fig:opinion_algo_chief}
\end{figure}

\paragraph{Precision \& recall diverge over time.}
We use the Supreme Court Data Base\footnote{We obtain data on features of cases by downloading case metadata from Washington University Law School's Supreme Court Data Base (SCDB).}, which contains metadata on SCOTUS cases, to see if any particular metadata can explain variation in summarization quality.
While we do not find notable variation across most of these features, we observe one exception: the divergence between recall and precision across all summaries increases over time.
Figure~\ref{fig:opinion_algo_chief} illustrates this trend, comparing summaries for opinions based on the Chief Justice of the Supreme Court at the time an opinion was issued. Summaries of earlier opinions, e.g., under the Warren Court, have greater parity between recall and precision compared to summaries from later opinions. One possible explanation for this trend is that opinions and syllabuses have become longer over time (Figure \ref{fig:dataset_stats} and Table \ref{tab:summary_describe}), while the summaries we evaluate show a growing disparity between their lengths and opinion/syllabus lengths over time.

\begin{figure*}[t]
    \centering
    \includegraphics[width=\linewidth]{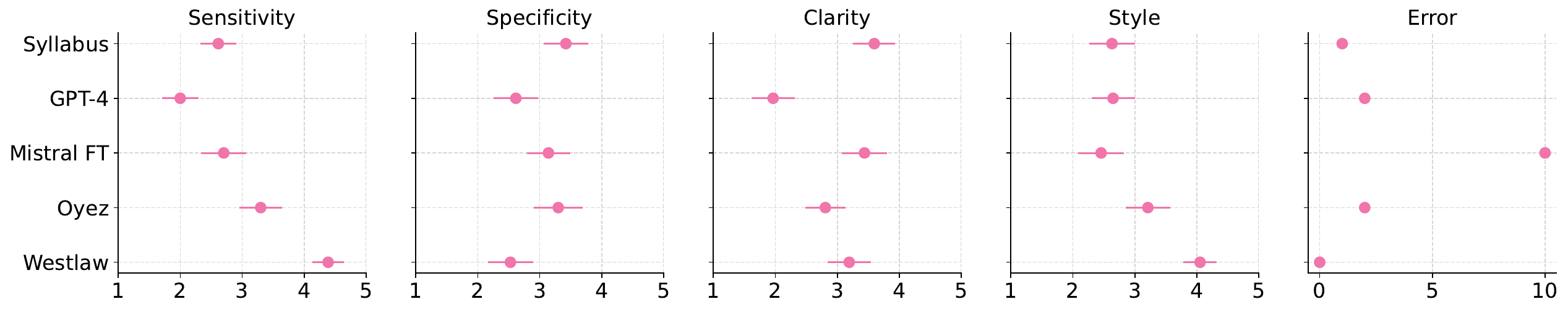}
    \caption{Human evaluation of model-generated and human summaries. x-axis is a rank, where 1 is best and 5 is worst. For \textbf{Error}, x-axis shows counts of the total number of errors identified by participants for each summary method. See Appendix \ref{sec:human-eval-dims} for explanation of each dimension.}
    \label{fig:opinion_human}
\end{figure*}

\input{assets/tables/auto-metrics/geval_human_all}

\subsection{Human Evaluation Disagrees with Automatic Evaluation}
\label{sec:human_results}

The results of our human evaluation, presented in Figure~\ref{fig:opinion_human} are distinctly different than those of our automatic evaluation. Whereas under automatic evaluation, Mistral FT outperforms other models as well as the control human-written summaries, we find that humans most commonly prefer GPT-4 summaries. GPT-4 particularly excels on \textit{clarity}, a crucial yet difficult to measure desideratum for the summarization task. Nonetheless, Mistral FT remains an above-average performer, successfully matching the original opinion syllabuses on every dimension except, importantly, number of errors. Evaluators report that roughly 20\% of Mistral FT summaries have at least 1 factual error, with a total of 10 errors identified across all evaluations. However, we see that these factual errors, or \textit{hallucinations}, are not necessarily a product of using LLMs, as GPT-4 has performance on par with syllabuses and Oyez in terms of factual correctness.

Surprisingly, the human evaluation also revealed that all three human-written summaries, including the official syllabuses, often performed worse than GPT-4. Westlaw summaries, despite being a paid service designed for legal professionals, ranked below average on sensitivity, clarity, and style. Even more intriguingly, the official syllabuses only matched or under-performed the LLM-generated summaries on all metrics except, crucially, factual correctness (\textit{error}). This result both challenges the assumption that human-written summaries are inherently superior, while also revealing opportunities and challenges in using LLMs for generating concise, correct, and accessible summaries.

\paragraph{LLM-based evaluation does not correlate better with human evaluations than traditional automatic metrics.}

\input{assets/tables/hallucinations}

The human correlation results in Table \ref{tab:auto_human_all_corr} illustrate differences in how well various evaluation strategies align with human preferences.

First, ROUGE and BERTScore metrics are more capable of capturing 
other aspects of summary quality, particularly sensitivity and style. 
ROUGE-L achieves the highest correlation with human judgements of style (0.47), and ROUGE-1/2/L outperform G-Eval on judgments of sensitivity. Despite its limitations, ROUGE offers useful signals for evaluating how well summaries balance inclusion and exclusion of content and how effectively they convey proper legal style.
It is worth noting that ROUGE shows the strongest negative correlation with specificity.

Second, G-Eval  shows significantly stronger correlations with human rankings of \textit{clarity} (Tables \ref{tab:default_geval_corr}, \ref{tab:adapted_geval_corr}) compared to BERTScore and ROUGE. This suggests that G-Eval better captures attributes like readability and logical flow, which are valued by human evaluators.

Third, G-Eval, while generally more consistent, performs similarly in its default and adapted versions, with only modest differences across dimensions. For instance, the adapted G-Eval slightly improves correlations with human judgements of style and specificity but shows no significant advantage for clarity or sensitivity. This suggests that while adapting prompts can impact G-Eval's results, it does not drastically alter its overall effectiveness.\footnote{In this analysis we focus on G-Eval's agreement with human judgments. Complete G-Eval scores are included in Appendix \ref{sec:geval-all} for reference.}

These findings highlight the need for evaluation metrics, whether LLM-based or not, that are more closely aligned with human preferences and capable of capturing granular dimensions, such as clarity, specificity, and style, that matter in human judgment of summaries.

\subsection{Error Analysis}
\label{sec:error}

\paragraph{Mistral hallucinates more conspicuously than GPT-4.}
We conduct further analysis of each summary flagged as containing factual errors according to the participants in the human evaluation. We compare each such summary to the original opinion to identify specific factual errors. Recent work has often referred to errors of this type as ``hallucinations''~\citep{huang2023surveyhallucinationlargelanguage}.

Table \ref{tab:hallucinations} presents example errors. Fine-tuned mistral contained the most errors in its summaries. Furthermore, these errors were more egregious than any produced in the GPT-4 Turbo summaries. These errors include simple factual errors (examples 1), incorrect citations (example 2), temporal understanding errors (example 3), as well as procedural history outcome errors (examples 4).

In contrast, GPT-4 Turbo errs in a more subtle way, failing to properly convey the legal analysis presented in the opinion (example 5) or misrepresenting background details (example 6). While the opinion indeed reverses the judgement of the court below, it does not reject its reasoning. Rather, the ruling is reversed due to a superseding issue of constitutionality. The summary generated by GPT-4 Turbo is thus incorrect in its characterization of the Supreme Courts decision.

\paragraph{Lexical variation.}
We define \textit{lexical variation} as the percentage of unique words in the summary not present in the opinion and consider it a measure of summary style. 
Mirroring our comparison of compression rates, syllabuses are shown to exhibit the lowest percentage of lexical variation from the original opinion. Surprisingly, the fine-tuned Mistral summaries have the highest average percentage of lexical variation at 41.7\%, even surpassing those written by Oyez (37.9\%). This is unexpected because Mistral FT is trained on legal syllabuses, while Oyez summaries are written for a general audience and might borrow less from the opinion. The high lexical variation rate of Mistral FT may be related to its higher rate of factual errors.

\begin{figure}[t]
\raggedright
\includegraphics[width=0.43\textwidth]{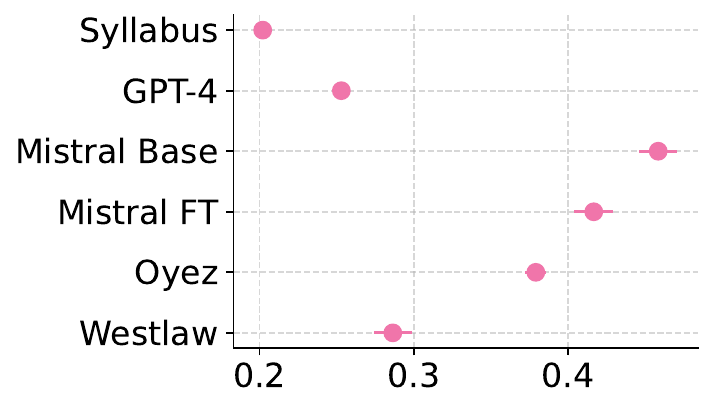}
\caption{Lexical Variation. Measures the fraction of words in summary that are not in opinion.}
\label{fig:novelty_compression}
\end{figure}

%% file: assets/tables/summary_describe.tex
\setlength\dashlinedash{0.2pt}
\setlength\dashlinegap{1.5pt}
\begin{table}
\centering\small
\begin{tabular}{l*{3}{c}}
\toprule
\textbf{Method}           &      \textbf{Length}& \textbf{Compression}& \textbf{Correlation} \\
\midrule
Opinion    &       6640  &   - &    - \\
\addlinespace[0.15em]
\hdashline\noalign{\vskip 0.5ex} 
\addlinespace[0.15em]
Syllabus    &       750 &       0.176&       0.676\textsuperscript{{\tiny\mbox{***}}} \\
GPT-4        &       321&       0.092&       0.088\textsuperscript{{\tiny\mbox{*}}} \\
Mistral Base     &       126&       0.034&       0.151\textsuperscript{{\tiny\mbox{***}}} \\
Mistral FT  &       447&       0.121&       0.179\textsuperscript{{\tiny\mbox{***}}} \\
\addlinespace[0.3em]
\hdashline\noalign{\vskip 0.5ex} 
Oyez        &       332&       0.096&       0.094\textsuperscript{{\tiny\mbox{*}}} \\
Westlaw     &       142&       0.044&       0.025 \\
\bottomrule
\end{tabular}
\caption{Descriptive statistics on summaries in test set ($n$=622). Length is number of words. Compression rate is ratio of words in syllabus to words in opinion.  Smaller number is more compression. Opinion included as reference. ($\textsuperscript{{\tiny\mbox{*}}} p<0.05$ $\textsuperscript{{\tiny\mbox{**}}} p<0.01$, $\textsuperscript{{\tiny\mbox{***}}} p<0.001$)}
\label{tab:summary_describe}
\end{table}

%% file: assets/tables/auto-metrics/geval_human_all.tex
\begin{table}[t]
\centering
\small

\begin{subtable}[t]{1\linewidth}
\centering
\resizebox{\textwidth}{!}{
\begin{tabular}{lrrrr}
\toprule
Metric & Sensitivity & Specificity & Clarity & Style \\
\midrule
ROUGE-1 & 0.54 & -0.28 & -0.02 & 0.40 \\
\addlinespace[0.3em]
ROUGE-2 & 0.55 & -0.32 & 0.02 & 0.39 \\
\addlinespace[0.3em]
ROUGE-L & 0.57 & -0.26 & 0.04 & 0.47 \\
\addlinespace[0.3em]
BERTScore & 0.45 & -0.11 & 0.01 & 0.43 \\
\bottomrule
\end{tabular}
}
\caption{ROUGE \& BERTScore correlations with human rankings.}
\label{tab:auto_human_corr}
\end{subtable}
 
\hfill

\begin{subtable}[t]{1\linewidth}
\centering
\resizebox{\textwidth}{!}{
\begin{tabular}{lrrrr}
\toprule
G-Eval & Sensitivity & Specificity & Clarity & Style \\
\midrule
Consistency & 0.26 & 0.06 & 0.25 & 0.11 \\
\addlinespace[0.3em]
Relevance & 0.45 & 0.09 & 0.26 & 0.30  \\
\addlinespace[0.3em]
Coherence & 0.42 & 0.05 & 0.28 & 0.33 \\
\addlinespace[0.3em]
Fluency & 0.43 & -0.15 & 0.08 & 0.30  \\
\bottomrule
\end{tabular}
}
\caption{Default G-Eval criteria and prompts used for GPT-4o-based evaluation.}
\label{tab:default_geval_corr}
\end{subtable}

\hfill

\begin{subtable}[t]{1\linewidth}
\centering
\resizebox{\textwidth}{!}{
\begin{tabular}{lrrrr}
\toprule
G-Eval & Sensitivity & Specificity & Clarity & Style \\
\midrule
Sensitivity & 0.43 & -0.01 & 0.19 & 0.23 \\
\addlinespace[0.3em]
Specificity & 0.51 & 0.06 & 0.24 & 0.38 \\
\addlinespace[0.3em]
Clarity & 0.38 & -0.02 & 0.23 & 0.21 \\
\addlinespace[0.3em]
Style & 0.42 & -0.07 & 0.08 & 0.34 \\
\bottomrule
\end{tabular}
}
\caption{Human evaluation criteria used for GPT4o-based evaluation. G-Eval criteria and prompts adapted accordingly.}
\label{tab:adapted_geval_corr}
\end{subtable}

\caption{Spearman correlations between various automatic scoring methods and human summary rankings. Interval scores from automatic methods are converted to rankings. Human rankings for the same set of summaries are averaged.}
\label{tab:auto_human_all_corr}
\end{table}

%% file: assets/tables/hallucinations.tex
\begin{table*}[t]
\centering
\small
\begin{tabular}{p{0.35\textwidth} p{0.65\textwidth}}
\toprule
\textbf{Hallucination in Summary} & \textbf{Explanation} \\ 
\midrule
\textit{Fine-tuned Mistral} & \\
\midrule
1. ``Petitioner, \textcolor{red}{a Negro}, applied for admission to the University of Washington Law School, a state-operated institution.'' & The opinion indicates the petitioner is not a member of a ``favored group'' nor a ``minority applicant''. This strongly implies \textcolor{blue}{the petitioner is white} and rules out the petitioner to be ``a negro''.  416 U.S. 312, 332, 325 (1974).  \\ 
\midrule 
2. ``\textcolor{red}{Sherbert v. Velelline, 416 U. S. 456}'' & The opinion cites ``\textcolor{blue}{Sherbert v. Verner, 374 U.S. 398}'' 494 U.S. 872, 875 (1990). \\ 
\midrule
3. ``He was held incommunicado for some five or seven days \textcolor{red}{after signing the statement}.'' & Petitioner Haynes testified he was held incommunicado until some five or seven days \textcolor{blue}{after his arrest}. 373 U.S. 503, 504 (1963). \\ 
\midrule
4. ``The District Court \textcolor{red}{ultimately entered judgment for petitioner}, holding that the Texas death penalty scheme was unconstitutional.'' & While the District Court initially stayed the execution pending judgement, it ultimately ``filed its findings and conclusions, \textcolor{blue}{rejecting each of the several grounds asserted by petitioner}. The writ was accordingly denied; also, the stay of petitioner's death sentence was vacated.'' 463 U.S. 880, 885 (1983). \\ 
\midrule
\textit{GPT-4 Turbo} & \\
\midrule
5. ``The Court \textcolor{red}{rejected the State's interest in [...] preserving the flag as an unalloyed symbol of the nation}'' & The Supreme Court did not reject the State's interest. They ``\textcolor{blue}{assume[d], arguendo, that it is [valid]}'' but found ``[t]he statute is nonetheless unconstitutional as applied to appellant's activity''. 418 U.S. 405, 414 (1974). \\ 
\midrule
6. ``The PCAOB is a regulatory body that \textcolor{red}{oversees the audits of public companies}'' & The PCAOB was created to govern \textcolor{blue}{the entire industry of accounting}, including ``hiring and professional development, promotion, supervision of audit work, the acceptance of new business and the continuation of old, internal inspection procedures, professional ethics rules, and `such other requirements as the Board may prescribe.'{''} 561 U.S. 477, 485 (2010). \\
\bottomrule
\end{tabular}
\caption{Comparison of model hallucinations and their explanations.}
\label{tab:hallucinations}
\end{table*}

%% file: sections/conclusion.tex
\section{Conclusion}
This paper introduces \dataset, a novel dataset for long-context summarization in the legal domain, comprising \casesummsize U.S. Supreme Court opinions and their official syllabuses. Our comprehensive evaluation of LLM-generated summaries, using both automatic metrics and expert human evaluation, reveals discrepancies between these assessment methods. While fine-tuned Mistral 7b outperforms larger models on automatic metrics, human experts rank GPT-4 summaries higher in clarity and accuracy. 
Our human evaluation also showed that GPT-4 summaries often outperformed human-written summaries, including official syllabuses and professional services, in several metrics except factual correctness.
LLM-based evaluation, such as G-Eval, may be a promising direction for reference-free evaluation but our results show that G-Eval does not correlate with human judgments better than traditional automatic metrics.
Our findings highlight the limitations of current automatic evaluation methods for legal summarization and underscore the importance of human evaluation in assessing summary quality, particularly in complex, high-stakes domains like law. This work contributes to the ongoing dialogue about evaluation methodologies in NLP and opens avenues for research in legal text summarization.

%% file: sections/limitations.tex
\section{Limitations}
First, the sample size of our human evaluation limits the conclusions we can draw.
Second, while we are able to offer insight into the value of fine-tuning, at least with respect to the open-source Mistral model, we are unable to estimate the value of prompt-engineering even the GPT-4 model because we do not have a natural benchmark, non-optimized prompt for that model.  A related weakness is that our evaluation of fine-tuning Mistral does not tell us the value of fine-tuning other models, such as GPT-4.  It is possible that the benefit to fine-tuning the latter may be lower than the former because GPT-4 is trained on more data and has far more estimated parameters. Third, we experiment with one LLM-based evaluation framework. While G-Eval is commonly cited and used, other LLM-based approaches could yield different results. Finally, while we demonstrate through a manual evaluation that our PDF extraction procedure is largely accurate (96\%), it is not perfect. A fraction of syllabuses, particularly those extracted from low-quality scans from SCOTUS opinions in the early 1800s, may not be fully correct.

%% file: sections/appendix.tex
\section{Summary Generation}
\label{sec:summary_gen_appendix}

\subsection{Mistral Fine-tuning and Generation Implementation Details}
\label{sec:implementation-details}

In our fine-tuning experiments, we use a batch size of 56. We select the best performing learning rate out of $\{2e-5, 2e-4, 2e- 3\}$ and early stop based on dev loss convergence. We conduct our experiments on 7 A100 80GB GPUs, with each fine-tuning run taking approximately 2 hours. During summary generation, we don't use sampling and set max tokens to 1500. We truncate opinions which exceed Mistral's 32768 context-length limit. In approximately 10\% of Mistral generations, the generation stops due to the length limit, rather than an \texttt{<eot>} token being generated. In such cases, we fallback to sampling a generation with \texttt{repetition\_penalty = 1.3} and \texttt{top\_p = 0.9}. This ensures a complete summary is produced and reduces degenerated summaries from the model.

\subsection{GPT-4 Generation Details}
To generate summaries based on majority opinions, we use the DSPy optimized prompt in Listing \ref{lst:gpt4_summarizaton_prompt_dspy}. The initial, unoptimized prompt, is included in Listing \ref{lst:gpt4_summarizaton_prompt}.

\noindent We run DSPy using \texttt{gpt-4-1106-preview}. We use the \texttt{SignatureOptimizer} (now called \texttt{COPRO}) with ROUGE-2 as the optimization metric along with a development set of reference syllabuses. Otherwise, we use default parameters.

\noindent For generating the final summaries after DSPy, we use \texttt{gpt-4-1106-preview} with \texttt{temperature = 0} and \texttt{max\_tokens = 1000}. All other parameters are set to the OpenAI API defaults.

\section{Automatic Evaluation}
\subsection{ROUGE Implementation Details}
We use the ROUGE implementation from the HuggingFace \texttt{evaluate} Python package. We set \texttt{use\_stemmer = True} and \texttt{use\_aggregator = True}. 

\subsection{BERTScore implementation Details}
We use the \texttt{bert-score} PyPI package. We use the default \texttt{bert-base-uncased} scoring model and all other default settings.

\subsection{BARTScore}
\label{sec:bartscore-results}
See results in Table \ref{tab:bartscore-eval-syllabus}.

\subsection{G-Eval LLM Evaluation}
\label{sec:geval-all}
\subsubsection{Generation Parameters}

We use use \texttt{gpt-4-0613} with \texttt{temperature = 1} and 
\texttt{n = 10}. All other parameters are set to the OpenAI API defaults.

\subsubsection{Default Prompts}
\textbf{Consistency}: see Listing \ref{lst:default_con}. \\
\textbf{Coherence}: see Listing \ref{lst:default_coh}. \\
\textbf{Relevance}: see Listing \ref{lst:default_rel}.\\
\textbf{Fluency}: see Listing \ref{lst:default_flu}.\\

\subsubsection{Adapted Prompts}
\textbf{Sensitivity}: see Listing \ref{lst:adapted_sens}.\\
\textbf{Specificity}: see Listing \ref{lst:adapted_spec}.\\
\textbf{Clarity}: see Listing \ref{lst:adapted_cla}.\\
\textbf{Style}: see Listing \ref{lst:adapted_sty}.\\

\subsubsection{Default \& Adapted G-Eval Scores}

Table \ref{tab:all-geval-scores} presents all the G-Eval scores.

\section{Human Evaluation}
\label{sec:human-eval-app}

\subsection{Dimensions of Summary Quality}
\label{sec:human-eval-dims}
\noindent\textbf{Sensitivity}: Does this summary include all relevant information required to understand the facts, judgment and reasoning? Outcome is a rank, where 1 is best, rank 5 is worst.  Ranks are mutually exclusive: only one case per rank. \vspace{0.5em} 

\noindent\textbf{Specificity}: Does this summary exclude irrelevant information that is not required to understand the facts, judgment and reasoning?  Rank from 1 to 5. \vspace{0.5em} 

\noindent\textbf{Clarity}: Is this summary clear and easy to read?  Rank from 1 to 5.\vspace{0.5em} 

\noindent\textbf{Style}: Does this summary have a legal style, defined as something written by a well-trained lawyer? Rank from 1 to 5.  For all measures where the outcome is rank, we mark the mean rank identically 3) with a red dashed line.\vspace{0.5em} 

\noindent\textbf{Factuality}: Does this summary contain any factual errors? (Yes/No). 

\subsection{Annotation Interface}
\begin{figure*}	
\begin{center}
    \includegraphics[width=\linewidth]{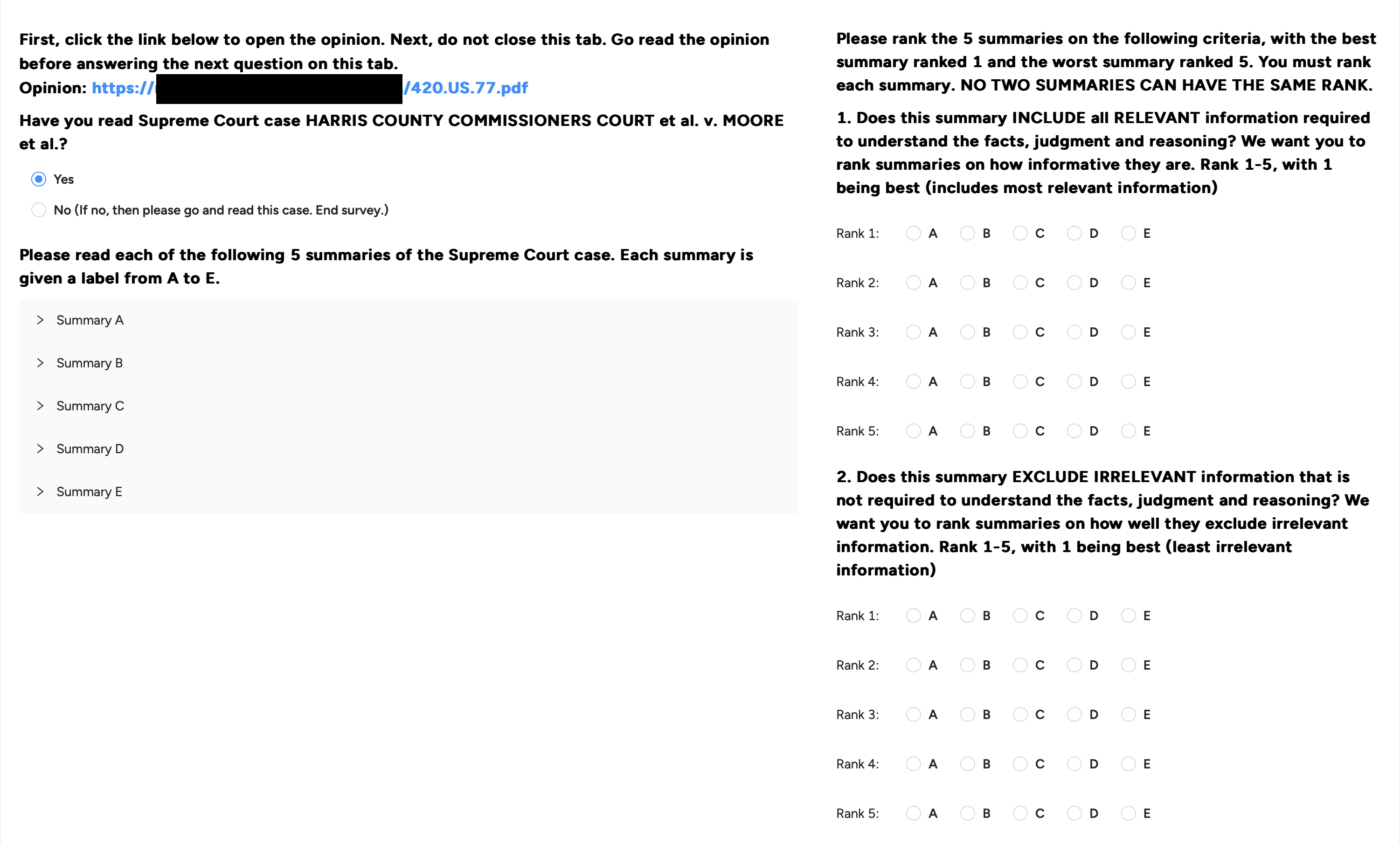}
    \includegraphics[width=\linewidth]{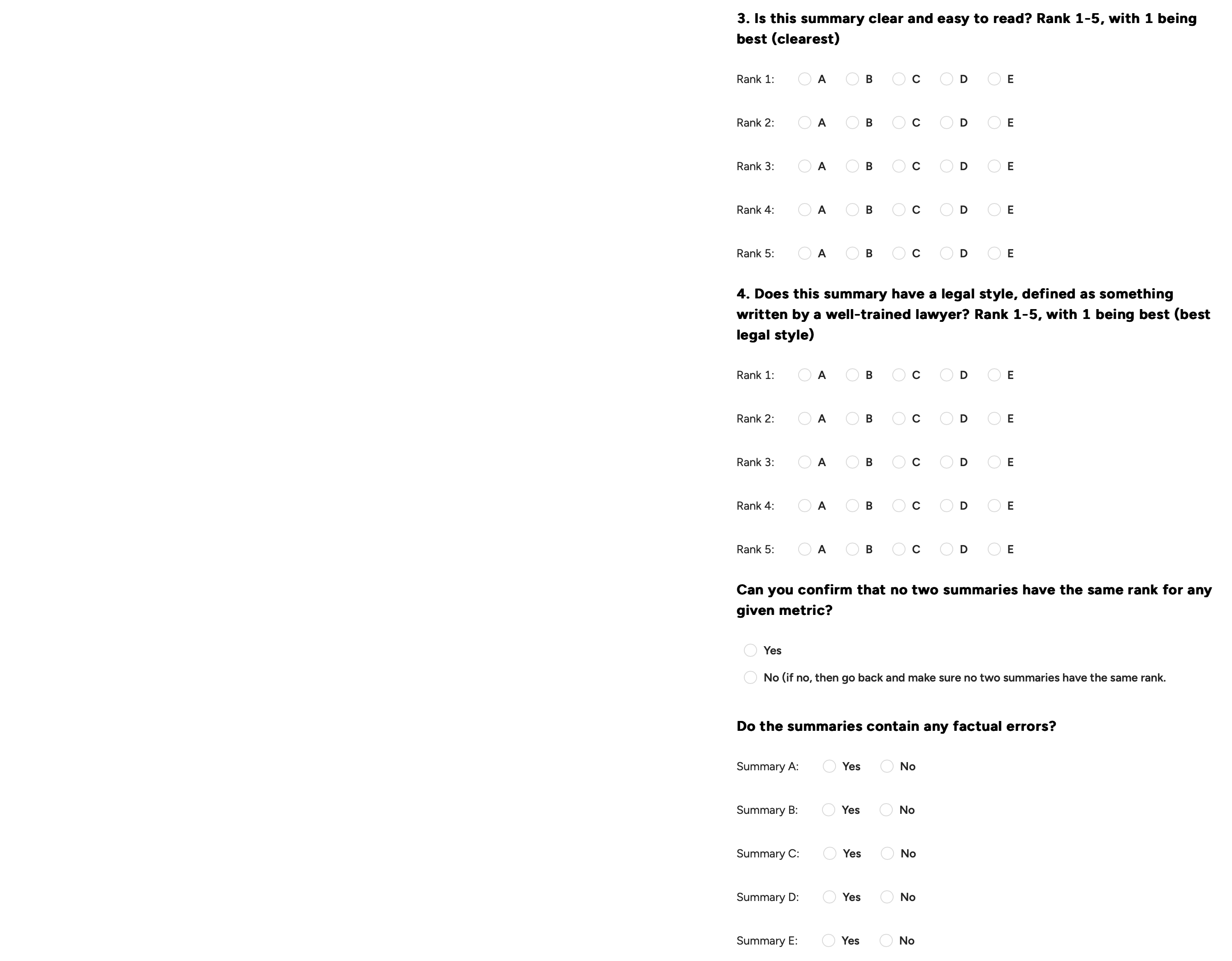}
    \caption{Labelstudio Annotation Interface}
    \label{fig:interface1}
\end{center}
\end{figure*}

Figure \ref{fig:interface1} is a screenshot of the annotation interface that participants used to read the opinions and summaries then rank them.

\subsection{Instructions \& Consent Materials for Participants}
\label{sec:consent}

\begin{figure*}
   \centering
   \includegraphics[width=0.7\hsize]{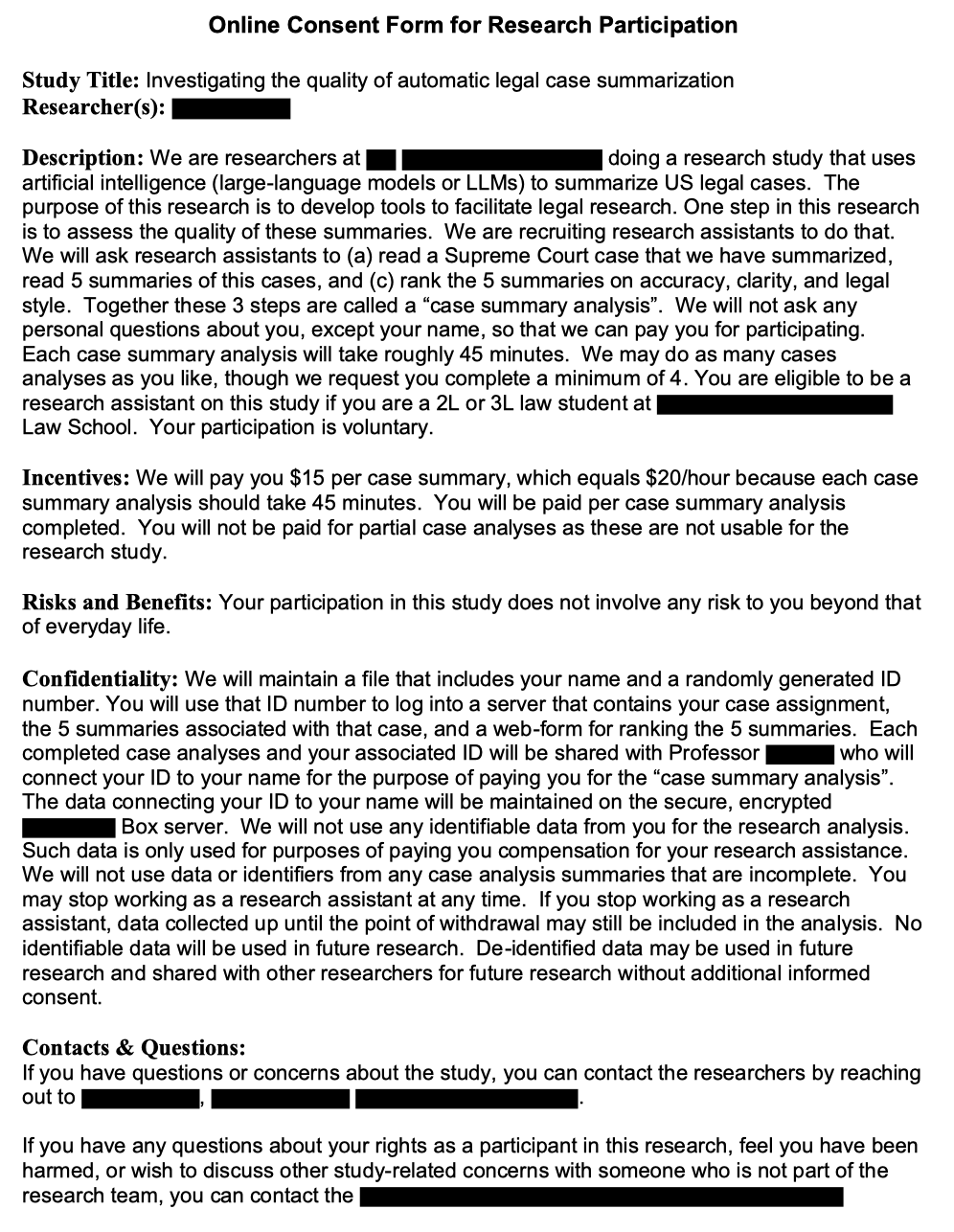}
   \includegraphics[width=0.7\hsize]{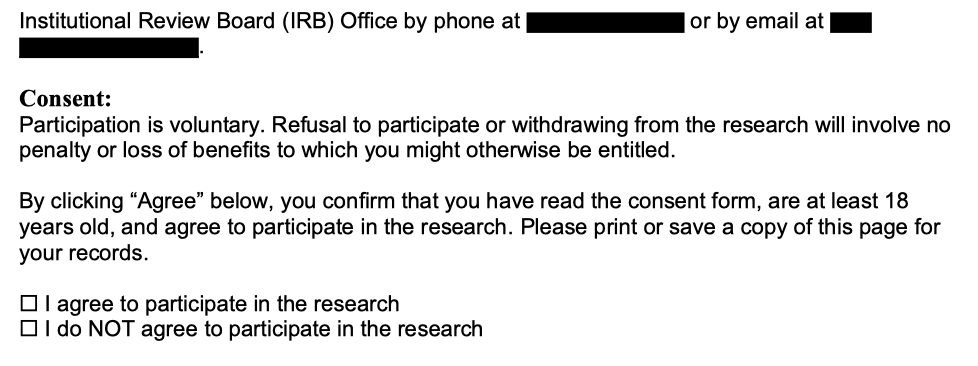}
   \caption{Consent form for research participation.}
    \label{fig:consent_form}
\end{figure*}

\begin{figure*}   
   \begin{center}
       \includegraphics[width=1\hsize]{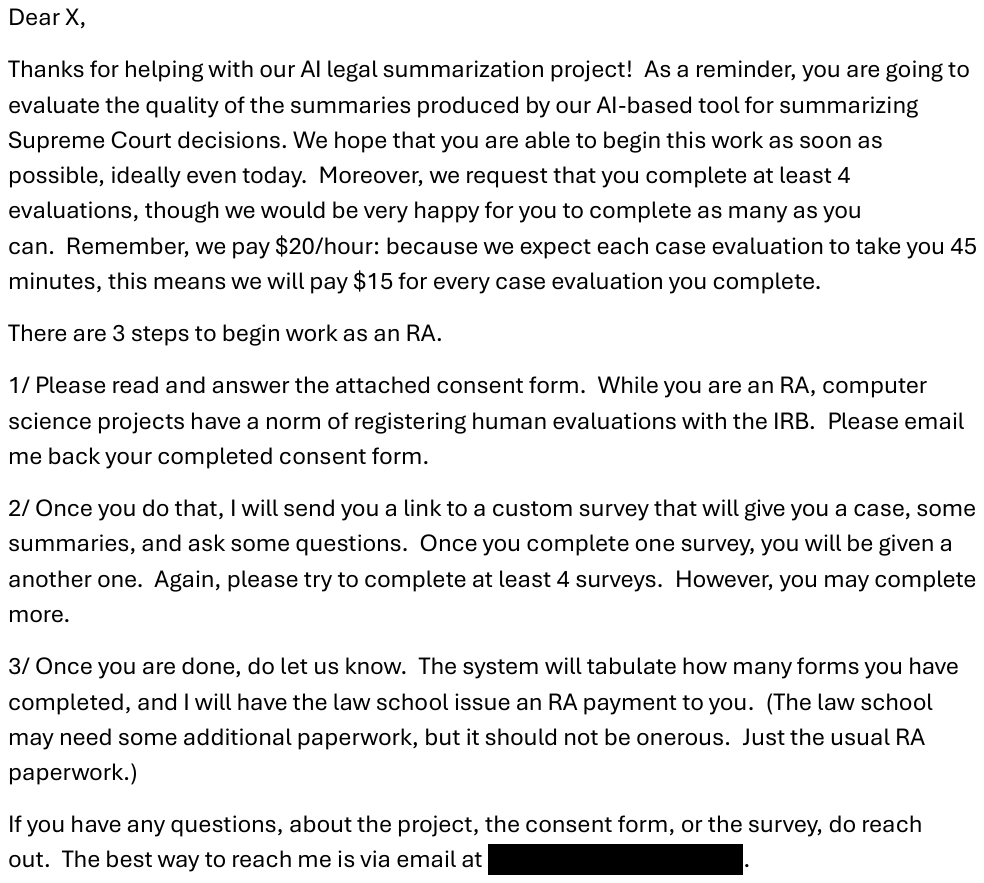}
       \caption{Email with instructions sent to participants.}
       \label{fig:participant_email}
   \end{center}
\end{figure*}

Figure \ref{fig:consent_form} shows the consent form presented to participants. \\Figure \ref{fig:participant_email} shows the email with annotation instructions sent to participants.

\begin{figure*}
\centering
\begin{lstlisting}[style=promptstyle, label={lst:gpt4_summarizaton_prompt}, caption={Initial GPT-4 Turbo summarization prompt used as input to DSPy.}]
Summarize the Supreme Court Opinion.

Opinion: {{OPINION}}
Summary:
\end{lstlisting}
\end{figure*}

\begin{figure*}
\centering
\begin{lstlisting}[style=promptstyle, label={lst:gpt4_summarizaton_prompt_dspy}, caption={DSPy optimized GPT-4 Turbo summarization prompt.}]
Review the provided Supreme Court opinion text. Deliver a concise, neutral summary that captures the essence of the legal reasoning, main points of law, conclusions drawn, and the implications of the decision, all whilst adhering to comprehensible language suitable for an educated general audience.

Opinion: {{OPINION}}

Summary of Supreme Court Opinion:
\end{lstlisting}
\end{figure*}

\input{assets/tables/auto-metrics/bartscores_syllabus}

\begin{figure*}
\centering
\begin{lstlisting}[style=promptstyle,, label={lst:default_con}, caption={Consistency prompt.}]
You will be given one summary written for a U.S. Supreme Court opinion.

Your task is to rate the summary on one metric.

Please make sure you read and understand these instructions carefully. Please keep this document open while reviewing, and refer to it as needed.

Evaluation Criteria:

Consistency (1-5) - the factual alignment between the summary and the summarized source. A factually consistent summary contains only statements that are entailed by the source document. Annotators were also asked to penalize summaries that contained hallucinated facts. 

Evaluation Steps:

1. Read the opinion carefully and identify the main facts and details it presents.
2. Read the summary and compare it to the opinion. Check if the summary contains any factual errors that are not supported by the opinion.
3. Assign a score for consistency based on the Evaluation Criteria.

Opinion Text:

{{Document}}

Summary: 

{{Summary}}

Evaluation Form (scores ONLY):

- Consistency:
\end{lstlisting}
\end{figure*}

\begin{figure*}
\centering
\begin{lstlisting}[style=promptstyle, label={lst:default_rel}, caption={Relevance prompt.}]
You will be given one summary written for a U.S. Supreme Court opinion.

Your task is to rate the summary on one metric.

Please make sure you read and understand these instructions carefully. Please keep this document open while reviewing, and refer to it as needed.

Evaluation Criteria:

Relevance (1-5) - selection of important content from the source. The summary should include only important information from the source document. Annotators were instructed to penalize summaries which contained redundancies and excess information.

Evaluation Steps:

1. Read the summary and the source document carefully.
2. Compare the summary to the source document and identify the main points of the opinion.
3. Assess how well the summary covers the main points of the opinion, and how much irrelevant or redundant information it contains.
4. Assign a relevance score from 1 to 5.

Opinion Text:

{{Document}}

Summary:

{{Summary}}

Evaluation Form (scores ONLY):

- Relevance:
\end{lstlisting}
\end{figure*}

\begin{figure*}
\centering
\begin{lstlisting}[style=promptstyle, label={lst:default_coh}, caption={Coherence prompt.}]
You will be given one summary written for a U.S. Supreme Court opinion.

Your task is to rate the summary on one metric.

Please make sure you read and understand these instructions carefully. Please keep this document open while reviewing, and refer to it as needed.

Evaluation Criteria:

Coherence (1-5) - the collective quality of all sentences. We align this dimension with the DUC quality question of structure and coherence whereby "the summary should be well-structured and well-organized. The summary should not just be a heap of related information, but should build from sentence to a coherent body of information about a topic."

Evaluation Steps:

1. Read the opinion carefully and identify the main topic and key points.
2. Read the summary and compare it to the opinion. Check if the summary covers the main topic and key points of the opinion, and if it presents them in a clear and logical order.
3. Assign a score for coherence on a scale of 1 to 5, where 1 is the lowest and 5 is the highest based on the Evaluation Criteria.

Opinion Text:

{{Document}}

Summary:

{{Summary}}

Evaluation Form (scores ONLY):

- Coherence:
\end{lstlisting}
\end{figure*}

\begin{figure*}
\centering
\begin{lstlisting}[style=promptstyle, label={lst:default_flu}, caption={Fluency prompt.}]
You will be given one summary written for a U.S. Supreme Court opinion.

Your task is to rate the summary on one metric.

Please make sure you read and understand these instructions carefully. Please keep this document open while reviewing, and refer to it as needed.

Evaluation Criteria:

Fluency (1-3): the quality of the summary in terms of grammar, spelling, punctuation, word choice, and sentence structure.

- 1: Poor. The summary has many errors that make it hard to understand or sound unnatural.
- 2: Fair. The summary has some errors that affect the clarity or smoothness of the text, but the main points are still comprehensible.
- 3: Good. The summary has few or no errors and is easy to read and follow.

Summary:

{{Summary}}

Evaluation Form (scores ONLY):

- Fluency:
\end{lstlisting}
\end{figure*}

\begin{figure*}
\centering
\begin{lstlisting}[style=promptstyle, label={lst:adapted_sens}, caption={Sensitivity prompt.}]
You will be given one summary written for a U.S. Supreme Court opinion.

Your task is to rate the summary on one metric.

Please make sure you read and understand these instructions carefully. Please keep this document open while reviewing, and refer to it as needed.

Evaluation Criteria:

Sensitivity (1-5) - the informativeness of the summary with respect to the opinion. An informative summary contains all relevant information required to understand the facts, judgement, and reasoning of the opinion.

Evaluation Steps:

1. Read the opinion carefully and identify the main facts, judgements, and reasoning it presents.
2. Read the summary and compare it to the opinion. Check if the summary misses relevant information presented in the opinion.
3. Assign a score for sensitivity based on the Evaluation Criteria.

Opinion Text: 

{{Document}}

Summary: 

{{Summary}}

Evaluation Form (scores ONLY):

- Sensitivity:
\end{lstlisting}
\end{figure*}

\begin{figure*}
\centering
\begin{lstlisting}[style=promptstyle, label={lst:adapted_spec}, caption={Specificity prompt.}]
You will be given one summary written for a U.S. Supreme Court opinion.

Your task is to rate the summary on one metric.

Please make sure you read and understand these instructions carefully. Please keep this document open while reviewing, and refer to it as needed.

Evaluation Criteria:

Specificity (1-5) - selection of important content from the source. The summary should include only important information from the source document and exclude irrelevant information that is not required to understand the facts, judgement, and reasoning. Annotators were instructed to penalize summaries which contained irrelevant and excess information.

Evaluation Steps:

1. Read the summary and the opinion carefully.
2. Compare the summary to the opinion and identify the main points of the opinion.
3. Assess how much irrelevant or redundant information it contains.
4. Assign a Specificity score from 1 to 5.

Opinion Text:

{{Document}}

Summary:

{{Summary}}

Evaluation Form (scores ONLY):

- Specificity:
\end{lstlisting}
\end{figure*}

\begin{figure*}
\centering
\begin{lstlisting}[style=promptstyle, label={lst:adapted_cla}, caption={Clarity prompt.}]
You will be given one summary written for a U.S. Supreme Court opinion.

Your task is to rate the summary on one metric.

Please make sure you read and understand these instructions carefully. Please keep this document open while reviewing, and refer to it as needed.

Evaluation Criteria:

Clarity (1-5): the quality of the summary in terms of grammar, word choice, and sentence structure. The summary should be well-structured and well-organized. The summary should not just be a heap of related information, but should build from sentence to a coherent body of information about a topic. Is this summary clear and easy to read? 

- 1: Very Poor. The summary is riddled with errors, making it very hard to understand. It may be disorganized to the point of confusion.
- 2: Poor. The summary has noticeable errors that impede clarity. It might be difficult to follow the central points or the flow of information.
- 3: Fair. The summary conveys the main points, but there are a few errors or awkward phrases. Overall, it is understandable but not polished.
- 4: Good. The summary is clear, well-structured, and mostly free of errors. The information is presented smoothly and cohesively.
- 5: Excellent. The summary is highly polished, with virtually no errors. It flows naturally, is easy to read, and effectively communicates the key information.

Summary:

{{Summary}}

Evaluation Form (scores ONLY):

- Clarity:
\end{lstlisting}
\end{figure*}

\begin{figure*}
\centering
\begin{lstlisting}[style=promptstyle, label={lst:adapted_sty}, caption={Style prompt.}]
You will be given one summary written for a U.S. Supreme Court opinion.

Your task is to rate the summary on one metric.

Please make sure you read and understand these instructions carefully. Please keep this document open while reviewing, and refer to it as needed.

Evaluation Criteria:

Style (1-3) - Does this summary have a legal style, defined as something written by a well-trained lawyer?

- 1: Poor. The summary does not read like something a trained lawyer would write. It may be overly casual, contain imprecise or incorrect legal terminology, or lack the logical structure and clarity typically found in legal writing.
- 2: Fair. The summary attempts a legal style and may use some appropriate terminology or reasoning, but it isn't fully polished. It might occasionally drift into non-legal language or lack the cohesive logic and precision expected from professional legal writing.
- 3: Good. The summary is stylistically consistent with something a trained lawyer would write. The language is precise, the reasoning is well-structured, and the tone is appropriately professional. It closely resembles the style commonly found in formal legal documents.

Evaluation Steps:

1. Read the summary and the opinion carefully.
2. Consider the style of the summary: Does it exhibit a tone, language, and structure similar to that of a well-trained lawyer (e.g., careful use of terminology, logical organization, professional tone)?
3. Compare the summary's style to the evaluation criteria. Assign a rating (1-3) based on how closely it aligns with a professional legal style, using the provided definitions for each level.

Opinion Text:

{{Document}}

Summary:

{{Summary}}

Evaluation Form (scores ONLY):

- Style:
\end{lstlisting}
\end{figure*}

\input{assets/tables/auto-metrics/all_geval_scores}

%% file: assets/tables/auto-metrics/bartscores_syllabus.tex
\setlength\dashlinedash{0.2pt}
\setlength\dashlinegap{1.5pt}
\begin{table*}
\centering
\begin{tabular}{@{}lccc@{}}
\toprule
   & \multicolumn{3}{c}{\textbf{BARTScore $\downarrow$}}  \\ 
 \cmidrule(lr){2-4}
\textbf{Method}   & P & R & F1  \\ \midrule
\addlinespace[0.3em]
GPT-4 Turbo         &  \textbf{256.0} & \underline{335.1} & \textbf{289.5} \\
\addlinespace[0.3em]
Mistral Base        & \underline{277.0} & 380.1 & 316.8 \\
\addlinespace[0.3em]
Mistral FT    & 297.9 & \textbf{312.3} & \underline{298.1} \\
\addlinespace[0.15em]
\hdashline\noalign{\vskip 0.5ex} 
\addlinespace[0.15em]
Oyez                 & 346.5 & \underline{334.6} & 339.3 \\
\addlinespace[0.3em]
Westlaw             & 307.8 & 345.2 & 323.5 \\
\bottomrule
\end{tabular}
\caption{BARTScores of model-generated and human-written summaries, where official syllabuses are the reference summaries. Sample includes 622 Supreme Court cases. There are 622 observations on each type of summary except Westlaw, for which we only have 156 observations. We report precision (P), recall (R), and F1-score (F1). BARTScores are negative log-likelihoods, so lower scores are better. We \textbf{bold} the best score(s) and \underline{underline} the second best score(s). For the scoring model, we use \href{https://huggingface.co/facebook/bart-large-cnn}{\texttt{facebook/bart-large-cnn}}, the default model used in \citet{yuan2021bartscore}
}
\label{tab:bartscore-eval-syllabus}
\end{table*}

%% file: assets/tables/auto-metrics/all_geval_scores.tex
\setlength\dashlinedash{0.2pt}
\setlength\dashlinegap{1.5pt}

\begin{table*}
\centering
\small
\resizebox{\textwidth}{!}{%
\begin{tabular}{@{}lcccc cccc@{}}
\toprule
\textbf{Method} & \multicolumn{4}{c}{\textbf{Default $\uparrow$}} & \multicolumn{4}{c}{\textbf{Adapted $\uparrow$}} \\ 

\cmidrule(lr){2-5} \cmidrule(lr){6-9}
  & \textbf{Consistency} & \textbf{Relevance} & \textbf{Coherence} & \textbf{Fluency} & \textbf{Sensitivity} & \textbf{Specificity} & \textbf{Clarity} & \textbf{Style} \\
\midrule
Syllabus  &  \underline{4.1} & \underline{3.6} & \underline{4.0} & \textbf{3.0} & \underline{3.3} & \underline{3.6} & 3.8 & \textbf{3.0} \\
\addlinespace[0.15em]
\hdashline\noalign{\vskip 0.5ex} 
\addlinespace[0.15em]
GPT-4 Turbo         & \textbf{4.4} & \textbf{4.2} & \textbf{4.5} & \textbf{3.0} & \textbf{3.9} & \textbf{3.8} & \textbf{4.3} & \textbf{3.0} \\
\addlinespace[0.3em]
Mistral FT  & 3.1 & 3.1 & 3.5 & \underline{2.8} & 2.8 & 3.0 & 3.8 & 2.6 \\
\addlinespace[0.15em]
\hdashline\noalign{\vskip 0.5ex} 
\addlinespace[0.15em]
Oyez        & 3.7 & 3.3 & 3.8 & \textbf{3.0} & 3.1 & 3.2 & \underline{4.0} & 2.8 \\
\addlinespace[0.3em]
Westlaw     & 3.8 & 3.2 & 3.6 & \textbf{3.0} & 3.0 & 3.0 & 3.8 & 2.7 \\
\bottomrule
\end{tabular}
}
\caption{G-Eval, default and adapted, LLM-based evaluation of model-generated and human-written summaries. Sample includes the 33 Supreme Court cases used for human evaluation. For each metric, we \textbf{bold} the best score(s) and \underline{underline} the second best score(s).
}
\label{tab:all-geval-scores}
\end{table*}